# How Part-of-Speech Tags Affect Text Retrieval and Filtering Performance

Robert M. Losee
Manning Hall, CB #3360
University of North Carolina
Chapel Hill, NC 27599-3360 U.S.A.

Phone: 919-962-7150
Fax: 919-962-8071
losee@ils.unc.edu

February 8, 1996

## Abstract

Natural language processing (NLP) applied to information retrieval (IR) and filtering problems may assign part-of-speech tags to terms and, more generally, modify queries and documents. Analytic models can predict the performance of a text filtering system as it incorporates changes suggested by NLP, allowing us to make precise statements about the average effect of NLP operations on IR. Here we provide a model of retrieval and tagging that allows us to both compute the performance change due to syntactic parsing and to allow us to understand *what* factors affect performance and *how*. In addition to a prediction of performance with tags, upper and lower bounds for retrieval performance are derived, giving the best and worst effects of including part-of-speech tags. Empirical grounds for selecting sets of tags are considered.

## 1 Introduction

Natural language processing contributes to improving retrieval performance in several ways (Lewis & Sparck-Jones, 1996). All of these must be reflected for retrieval or filtering purposes in either a modified query or in a modified document. We present a model of retrieval and part-of-speech tagging that allows us to make specific claims about the level of performance that will be obtained with such linguistically induced modifications, including tagging, and provides both upper and lower bounds for performance with the best-case and worst-case tagging performance. The analytic modeling of query expansion and the incorporation of term dependencies and relevance feedback was examined earlier by Losee (1995, 1996a), which serve as companion articles to this work.

Most research examining either linguistic phenomenon or information retrieval and filterings systems, or both, describe system performance by examining statistical, average performance figures, or the characteristics of specific instances or phenomena. However, exact expected results may be obtained in some instances through modeling enough aspects of a filtering system to provide exact statements about the expected characteristics of the system. This is different than computing a single average for an experimental data set. The analytic approach allows us to formally understand how variables interact to produce a certain level of performance and thus complementing experimental methods. While the impact of some assumptions, as well as some models of natural language processing, may best be studied empirically, others questions are best answered analytically.

Tags from a set of commonly accepted grammatical constructs are usually assigned in a computational environment though the *tagging* process (Brill, 1994). Assigning a part-of-speech tag to a token allows users to discriminate between different senses in which a term is used. A part-of-speech tag assigned to the term *girl* in sentences like *girl bites dog* and *dog bites girl* can be used by a retrieval system to retrieve only those documents in which a girl acts as biter (subject) or those documents in which the girl is bitten (object). While using natural language processing techniques such as tagging may improve retrieval and filtering performance, the degree of improvement varies from minimal to moderately helpful (Burgin & Dillon, 1992; Strzalowski, 1995). The author hopes that the analytic work described below will lead to an understanding of how tagging contributes toward the matching of queries and text.

We make several simplifying assumptions here to allow us to understand the nature and benefits of part-of-speech tagging. For example, we limit ourselves to queries with a single term to allow us to examine the impact of tagging on a single term, taken in isolation. We similarly assume optimal retrieval so that we can avoid the complexity added by working with a suboptimal ranking method. Earlier work on analytic prediction of retrieval performance shows how a model may assume multiple terms and suboptimal retrieval, but such complex models may hide some simple underlying phenomena (Losee, 1995, 1996a).

## 2 An Analytic Model of Text Retrieval

Retrieval and filtering systems may be evaluated using any of a number of measures, described in standard retrieval texts (Salton & McGill, 1983; Van Rijsbergen, 1979). Most of these retrieval measures capture different aspects of the placement of documents considered "relevant" by the user near the front of an ordered list of documents, with non-relevant documents being moved toward the bottom of the list. We believe that these measures are also adequate as indicators of the performance of document filters as well as text retrieval performance. This work uses the Average Search Length (ASL), the average number of documents or text fragments examined in moving down a ranked list of documents until arriving at the average position of a relevant document. This number is easily interpreted and, more importantly, is far better suited as a measure to be predicted than are most other retrieval and filtering measures. A worked-out example is given at the end of this section.

The analytic model of retrieval suggests that optimal document ranking is the ranking of documents by the probability they are relevant, given the characteristics of the document (Losee, 1994). We limit ourselves here to queries with a single term, with the multi-term case being extrapolated from this single term model. Note that
$$\Pr(rel|d) = \frac{\Pr(d|rel)\Pr(rel)}{\Pr(d)},$$
where $\Pr(d|rel)$ denotes the probability a document has the query term with binary frequency $d = 1$, indicating the term is present in the document, given that the document is relevant, and where $\Pr(d)$ denotes the unconditional probability a document has the query term with binary frequency $d = 1$.

The documents are ranked for presentation to the user, who continues to retrieve documents until deciding to stop. Let us assume that documents with the characteristic $(d)$ are ranked ahead of those without the characteristic $(\overline{d})$, and where $\Pr(\overline{d}) = 1 - \Pr(d)$. The midpoint for those ranked documents with the query feature, $d$, is $N\Pr(d)/2 + 1/2$. Thus, if we had 10 documents, half with the term, the mid point for those five documents with the term would be at document number $10(.5/2) + 1/2 = 3$. The midpoint for the portion of those documents at the bottom end of the ranking, those with frequency of $\overline{d}$, is $N(1 - \Pr(\overline{d})/2) + 1/2$. The Average Search Length thus may be computed as

$$ASL = N[\Pr(d|rel)\Pr(d)/2 + \Pr(\overline{d}|Rel)(1 - \Pr(\overline{d})/2)] + 1/2. \qquad (1)$$

For notational simplicity below, we will denote $t = \Pr(d)$ and $p = \Pr(d|rel)$. If we assume that retrieval is optimal and we continue with a single term, the ASL

Table 1: Sample documents, where "Y" denotes *yes* and "N" denotes *no*.

| *Relevant Documents* | | *Non-relevant Documents* | |
| --- | --- | --- | --- |
| *Term Present?* | *Query Tag?* | *Term Present?* | *Query Tag?* |
| Y | Y | Y | N |
| Y | Y | Y | Y |
| Y | N | N | N |
| N | N | N | N |
| N | N | N | N |

may be computed as
$$ASL = \frac{N}{2}(1 + t - p) + \frac{1}{2}, \qquad (2)$$
a reformulation of Equation 1. This combines the middle position in the ranked list of documents times a factor that increases (decreases performance) as the query term increases in relative frequency and decreases the ASL (increases performance) as the query term increases in relative frequency in the relevant documents, plus a constant. Below we concentrate on the middle of Equation 2, which we refer to as the $A$ factor,
$$A = 1 + t - p. \qquad (3)$$
The $A$ factor is the query dependent portion of the formula that computes ASL. When $A$ is above 1, performance is worse than that obtained with random retrieval, while when $A$ is below 1, performance is better than that obtained with random retrieval.

A set of ten sample documents is provided in Table 1 for a query term that has been grammatically tagged. This data has the following parameter values: $t = 1/2$ and $p = 3/5$. For all examples for the rest of this paper we assume that $N = 10$. We find that when we wish to estimate the ASL without using tagging information, the A component is $1 + 1/2 - 3/5 = 9/10$. The ASL may be computed as $10/2(9/10) + 1/2 = 5$.

## 3   Part-of-Speech Tags and IR Performance

When tagging documents, a smaller or the same number of documents will have the term tagged as in the query after tagging than have the untagged term before tagging takes place. We denote the probability that a document is tagged with the query tag, given that it has term $d$, as $\tau$. The probability that a document has the term and is tagged with the query tag is the product $t\tau$. Similarly, the probability that a term is tagged with the query tag, given that it is in a relevant document and has the term, is $\pi$. The probability that a document has the tagged term and is relevant is the product $p\pi$.

A term is assigned a part-of-speech tag with the expectation that the tagging will increase retrieval or filtering performance. Filtering performance is improved if and only if the ASL with the tagging is less than the ASL without the tagging, or

$$1 + t\tau - p\pi < 1 + t - p. \tag{4}$$

We can measure the degree to which retrieval performance with tagging exceeds the performance without tagging if Equation 4 is reformulated so that $c$ represents the amount added to the left hand side of Equation 4 by tagging, with a positive value for $c$ indicating that tagging improves performance (decreases ASL), and a negative value indicates that tagging decreases system performance. We refer to $c$ as the *Tagging Improvement Factor* (TIF).

$$1 + t\tau - p\pi + c = 1 + t - p \tag{5}$$
$$c = t(1 - \tau) - p(1 - \pi) \tag{6}$$

The TIF $c$ is thus computed as the proportion of all documents with the term in question that aren't tagged minus the proportion of relevant documents with the term in question that aren't tagged. A TIF of $c$ is 0 when the same proportion of all documents as relevant documents are assigned the tag in question. This occurs when the tags are distributed the same in both the set of relevant documents and in all documents. If a tag occurs with greater relative frequency in the relevant documents, the tagging results in improved performance.

The data in Figures 1 and 2 show the break-even points between text filtering performance with and without part-of-speech tags, that is, where the TIF ($c$) is 0. Giving varying values of $p$, $\tau$, and $\pi$, and $t = 1/10$ for Figure 1 and $t = 1/2$ for Figure 2, we find that, for lower values of $t$, the $\pi$ value rises to both higher values and rises more quickly than for higher values of $t$. This suggests that tagging is more likely to have potential benefit (having a lower break-even point) for higher values of $t$. This is different than the simple conclusion that one could derive from Equation 2, which suggests that the performance drops as $t$ or $\tau$ increases.

When tagging is applied to the data in Table 1, where $\tau = 1/2$ and $\pi = 2/3$, we compute the ASL as $\frac{10}{2}(1 + \frac{1}{2}\frac{1}{2} - \frac{2}{3}\frac{3}{5}) + \frac{1}{2} = 4.75$. If one examines Figure 2 one finds that for $t = 1/2$, $p = 3/5$, $\pi = 2/3$ and $\tau = 1/2$, we are at a point above the break-even surface but not very far from the surface.

Figure 3 shows the ASL when $t = 1/2$ and $p = 3/5$ for both a tagged query (the fine mesh) and for an untagged query (with larger holes in the mesh). Figure 4 similarly shows the ASL when $t = 1/10$, with everything else similar except for the ASL. Note that for the case where there is a low $t$ and a large gap between $t$ and $p$, the terms are already good discriminators and the region where tagging will improve performance (the top part of the figure where the large mesh is above the finer mesh, the latter representing tagged performance) is small. For this data, tagging results in improved performance with very high $\pi$ and, to a lesser extent, for lower $\tau$.

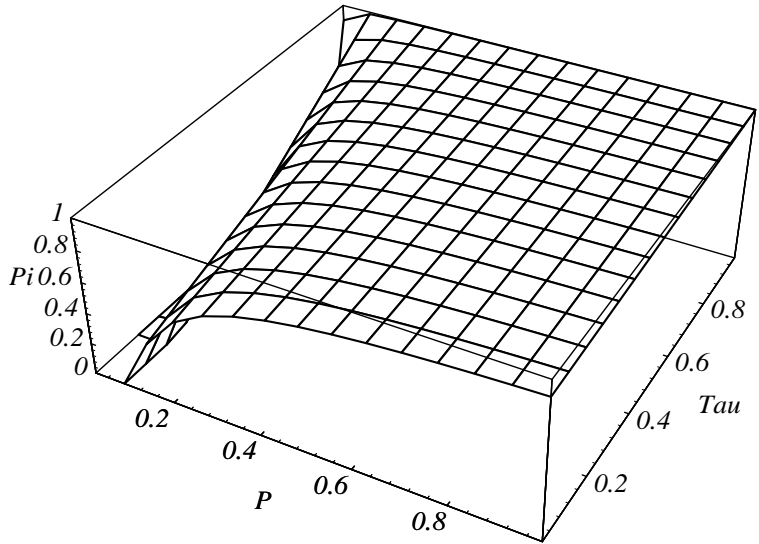

Figure 1: The break-even $\pi$ ("pi") points for deciding to tag or not tag a term, with $t = 1/10$.

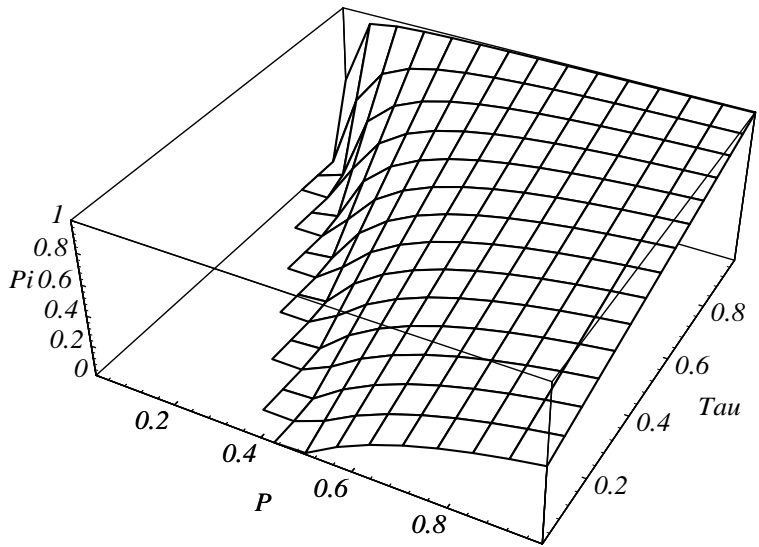

Figure 2: The break-even $\pi$ ("pi") points for deciding to tag or not tag a term, with $t = 1/2$.

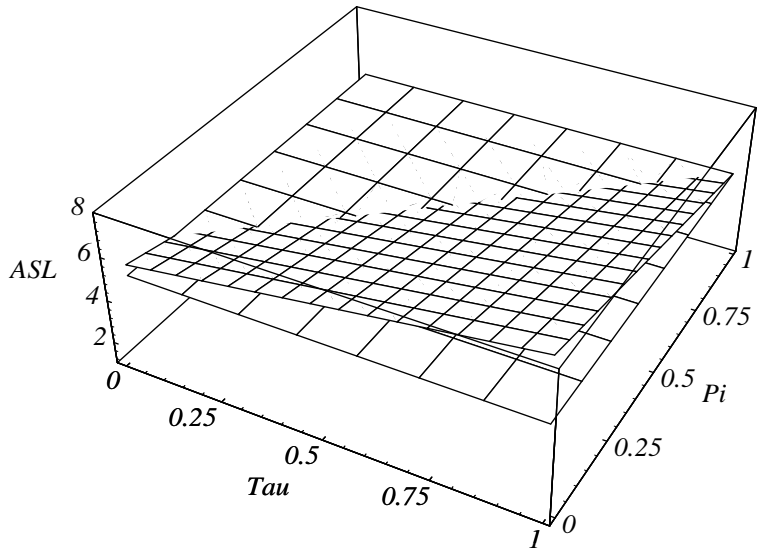

Figure 3: The Average Search Lengths (ASL) for two searches. The large mesh (large holes) represents retrieval with no tags and with $t = 1/2$ and $p = 3/5$. The smaller mesh represents a search with tagging, where the parameters are as above but where $\pi$ ("Pi") and $\tau$ ("Tau") are allowed to vary.

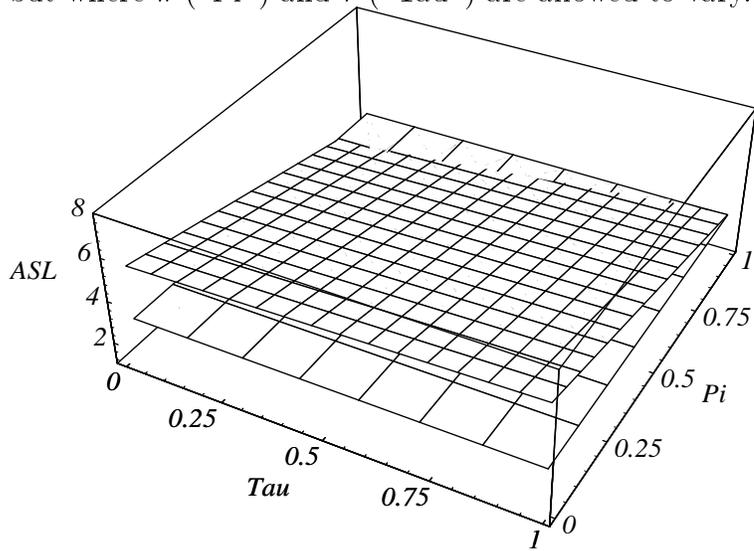

Figure 4: The Average Search Lengths (ASL) for two searches. The large mesh (large holes) represents retrieval with no tags and with $t = 1/10$ and $p = 3/5$. The smaller mesh represents a search with tagging, where the parameters are as above but where $\pi$ ("Pi") and $\tau$ ("Tau") are allowed to vary.

## 4 Best and Worst-case Performance with Tagging

Retrieval performance may be viewed as dominated by $t$ and $p$. Part-of-speech tagging may be viewed as providing a means of improving on this by modifying these parameters through $\tau$ and $\pi$, but it cannot surmount some limits imposed by the untagged probabilities. The highest TIF (Equation 6) occurs when the proportion of all documents with the query term tagged as in the query approaches its minimum ($\tau \to 0$) and the proportion of relevant documents with the query term tagged as in the query approaches its maximum ($\pi \to 1$). The worst case occurs when $\tau \to 1$ and $\pi \to 0$. The bounds for retrieval performance are thus

$$\text{W}orst\ case \qquad\qquad \text{B}est\ case \qquad\qquad (7)$$
$$\frac{N}{2}(1+1t-0p)+1/2 \geq ASL \geq \frac{N}{2}(1+0t-1p)+1/2$$
$$\frac{N}{2}(1+t)+\frac{1}{2} \geq ASL \geq \frac{N}{2}(1-p)+\frac{1}{2}\ \text{a}nd \qquad (8)$$
$$1+t \geq A \geq 1-p \qquad (9)$$

We can see that performance will be proportionally no better than would be obtained with an $A$ factor of $1-p$ and no worse than $1+t$. The range of retrieval performance is thus bracketed between that obtainable with optimal tagging on the right hand side of Equation 8 and that with the worst case tagging on the left hand side.

In reality, $\tau$ doesn't approach very close to 0 with small databases, such as in Table 1, although $\tau$ does approach 0 with larger databases. With smaller databases, or when it is desirable to compute the exact value, it becomes necessary to take into account when computing the best-case and worst-case performance values the failure of the limits to be met (by a factor $rp$, where $r = \Pr(rel)$). The exact bounds are thus:

$$\frac{N}{2}(1+t-rp)+\frac{1}{2} \geq ASL \geq \frac{N}{2}(1+rp-p)+\frac{1}{2} \qquad (10)$$

As the size of the database grows, and $t$ approaches 0, then Equation 8 approaches the correct values. Note that the $rp$ values in Equation 10 become very small, approaching 0 when a very large, realistic database is used, so that this equation in the limiting case approaches Equation 8.

In most realistic searches, $t$ will be rather small if it is a good search term, usually below .01. When $p$ is much higher than this, as would be the case with a strongly discriminating term, the potential for improvement with tagging is far better than the potential decrease in performance.

Using the data in Table 1, $N = 10$, $p = 3/5$ and $t = 1/2$, and using the best and worst case measures described in Equation 8, $8 \geq ASL \geq 2.5$. Using the exact formula (Equation 10), however, we find the $6.5 \geq ASL \geq 4$. The ASL computed earlier for this tagged query was 4.75. It is clear that this is better than the ASL of 5 found when no tagging occurs but is not as good as it could be, not equaling the best-case performance (4).

Optimal or best-case tagging and the ASL of 4 could be obtained here by tagging all relevant documents with the term in Table 1 and by not tagging those non-relevant documents with the term. In this situation, the three relevant documents with the term would be at the beginning of the ranked list (average position 2) and the other two relevant documents would be at the average position of 7, producing $ASL = 4$. The worst-case value of 6.5 is obtained when the 2 non-relevant documents with the term are tagged, and all the other documents are untagged, with the position of these untagged documents having a center position of 6.5.

## 5 Discussion and Conclusions

Analytic models of text filtering and retrieval can be used to determine the magnitude and direction of change that occurs when various forms of natural language processing are applied to queries and documents. In earlier work, the author examined the analytic modeling of performance using query revision, through relevance feedback, and through query expansion (Losee, 1995, 1996a). In the work above, we extended this work to model the effect of part-of-speech tagging on filtering performance.

Part-of-speech tags may be assigned to terms in documents and queries, providing additional information about the similarities and differences between the documents and queries. The performance of a retrieval system may be estimated analytically and the performance with and without part-of-speech tags may be compared. We can see that as the probability that a query term is tagged if it is in a relevant document increases, the ASL decreases. We may also compute the break-even points for tagging (setting the Term Improvement Factor $c$ in Equation 6 to 0, as shown in Figures 1 and 2), providing a tool that will help us decide whether a tagging system is likely to improve performance. The best-case and worst-case performance after the assignment of tags is described by Equation 8.

Further work may examine the empirical basis for tags (Yngve, 1986). This problem was explored by Losee (1996b) who used genetic algorithms to develop part-of-speech tags. While there is an intellectual basis for the tags most commonly used, a formal empirical basis for choosing a particular set of tags would be desirable. We believe that the pragmatic measure of tags' utility provides such a basis. The ideal set of tags is that set which results in the maximum average TIF for terms in documents and in queries for a particular set of queries, documents, and relevance judgments. One might, for example, examine the relationship between "traditional" tags, based on traditional linguistic categories, compared to empirically supported tags derived from a system computing Tag Improvement Factors for the different tags. An understanding of the foundation of tagging may serve as the basis for learning sets of tags by either automated systems or by children learning language.